\documentclass{article}

\usepackage{microtype}
\usepackage{graphicx}
\usepackage{subfigure}
\usepackage{booktabs} % for professional tables

\usepackage{hyperref}

\usepackage[accepted]{ICML2024/icml2024}

\usepackage{amsmath}
\usepackage{amssymb}
\usepackage{mathtools}
\usepackage{amsthm}

\usepackage[capitalize,noabbrev]{cleveref}

\theoremstyle{plain}
\newtheorem{theorem}{Theorem}[section]

\theoremstyle{definition}

\theoremstyle{remark}
\newtheorem{remark}[theorem]{Remark}

\usepackage[textsize=tiny]{todonotes}

\icmltitlerunning{Geometry-Aware Normalizing Wasserstein Flows for Optimal Causal Inference}

\usepackage{amsmath,amsfonts,bm}

\def\1{\bm{1}}

\def\mI{{\bm{I}}}

\def\mT{{\bm{T}}}

\def\mX{{\bm{X}}}

\DeclareMathAlphabet{\mathsfit}{\encodingdefault}{\sfdefault}{m}{sl}
\SetMathAlphabet{\mathsfit}{bold}{\encodingdefault}{\sfdefault}{bx}{n}

\def\gF{{\mathcal{F}}}

\def\gM{{\mathcal{M}}}

\def\gP{{\mathcal{P}}}

\def\gT{{\mathcal{T}}}

\def\gW{{\mathcal{W}}}

\newcommand{\E}{\mathbb{E}}

\newcommand{\R}{\mathbb{R}}

\newcommand{\Var}{\mathrm{Var}}

\DeclareMathOperator*{\argmin}{arg\,min}

\DeclareMathOperator{\Tr}{Tr}

\begin{document}

\twocolumn[
\icmltitle{Geometry-Aware Normalizing Wasserstein Flows for Optimal Causal Inference}



\icmlsetsymbol{equal}{*}

\begin{icmlauthorlist}
\icmlauthor{Kaiwen Hou}{sch}
\end{icmlauthorlist}

\icmlaffiliation{sch}{Columbia Business School, New York, NY}

\icmlcorrespondingauthor{Kaiwen Hou}{KHou24@gsb.columbia.edu}
%\icmlcorrespondingauthor{Firstname2 Lastname2}{first2.last2@www.uk}

\icmlkeywords{Machine Learning, ICML}

\vskip 0.3in
]



\printAffiliationsAndNotice{\icmlEqualContribution} % otherwise use the standard text.

\begin{abstract}
This paper presents a groundbreaking approach to causal inference by integrating continuous normalizing flows (CNFs) with parametric submodels, enhancing their geometric sensitivity and improving upon traditional Targeted Maximum Likelihood Estimation (TMLE). Our method employs CNFs to refine TMLE, optimizing the Cramér-Rao bound and transitioning from a predefined distribution $p_0$ to a data-driven distribution $p_1$. We innovate further by embedding Wasserstein gradient flows within Fokker-Planck equations, thus imposing geometric structures that boost the robustness of CNFs, particularly in optimal transport theory.

Our approach addresses the disparity between sample and population distributions, a critical factor in parameter estimation bias. We leverage optimal transport and Wasserstein gradient flows to develop causal inference methodologies with minimal variance in finite-sample settings, outperforming traditional methods like TMLE and AIPW. This novel framework, centered on Wasserstein gradient flows, minimizes variance in efficient influence functions under distribution $p_t$. Preliminary experiments showcase our method's superiority, yielding lower mean-squared errors compared to standard flows, thereby demonstrating the potential of geometry-aware normalizing Wasserstein flows in advancing statistical modeling and inference.
\end{abstract}

\section{INTRODUCTION}

The fidelity of causal effect estimates is substantially influenced by the underlying statistical models in the field of causal inference. TMLE, a pivotal methodology in this domain, utilizes a series of parametric submodels to converge towards the true data-generating process. Nonetheless, the efficiency of TMLE hinges on the precise definition of the perturbation direction, which in turn is contingent on the accuracy of the specified propensity models—a requirement often strenuous to fulfill in practice.

Our contribution lies in the application of continuous normalizing flows (CNFs) to enrich the geometric delineation of these parametric submodels within the TMLE framework. CNFs, with their capacity for complex distribution modeling through a differential equation framework, introduce a heightened level of flexibility and geometric awareness to the estimation process. Specifically, our technique contrasts with traditional TMLE by embarking on a path that minimizes the Cramér-Rao bound from a priori knowledge encapsulated in $p_0$ towards an empirically-informed distribution $p_1$.

We extend the typical use of CNFs, which often relies on Fokker-Planck equations to describe probabilistic transitions, by employing Wasserstein gradient flows. This empowers a more efficient and acute navigation through the space of models, attentive to the evolution of density functions.

Drawing from the theoretical work of \citet{chen2018neural}, we assert that while Fokker-Planck equations provide a robust theoretical foundation for such transformations, the incorporation of Wasserstein gradient flows caters to a more functional and specific implementation in the context of causal inference. This perspective allows us to bypass the limitations identified by \citet{liu2022neural} and \citet{raissi2019physics} related to the parametrization of these equations, fostering a CNFs utilization that is more informed and structurally sound. Consequently, our approach enables an estimation that is both nuanced and geometrically congruent, aligning well with the intricate requirements of causal inference.

Furthermore, the versatility of CNFs permits the imposition of additional structures or criteria based on prior objectives. If the true distribution is suspected to reside within a particular manifold of statistical submodels, our methodology can constrain the flow transformations to honor this geometric structure. This approach transcends regularization, becoming a pivotal element for enhancing model robustness, such as by integrating optimal transport theory into the transformations.

Acknowledging the numerical tractability of instantaneous changes of variables in CNFs, as noted by \citet{chen2018neural}, we also address the challenge presented by neural network parameterizations of the Fokker-Planck equations, which traditionally depend on sampling methods. Inspired by the innovation in physics-informed neural networks \citep{raissi2019physics}, our paper advocates for a sophisticated embedding of the functional forms of the PDE within CNFs, bolstering their application in causal analysis.

\section{PRELIMINARIES}

This section presents an overview of the general formulation of Wasserstein gradient flows, detailing their interrelation with continuous normalizing flows.

\subsection{Wasserstein metric}
Consider a Radon space~$(\gP_2(\Omega),\gW_2)$ where $\mathcal{P}_2(\Omega)$ denotes the set of $L_2$-probability measures on a compact sample space~$\Omega$, and $\gW_2$ represents the 2-Wasserstein metric
\begin{align*}
    \gW_2^2(\mu,\nu) = \inf_{\gamma\in\Gamma(\mu,\nu)} \int_{\gP_2(\Omega)\times \gP_2(\Omega)} \|x_1-x_2\|^2 \, d\gamma(x_1,x_2).
\end{align*}
This metric is defined for any pair of probability measures $\mu$ and $\nu$, both possessing finite second moments, and it involves the set $\Gamma(\mu, \nu)$, which encompasses all possible couplings of $\mu$ and $\nu$. 
Similarly defined $q$-Wasserstein metric for $q\geq 1$ is also a well-defined distance on $\gP_q(\Omega)$~\citep{santambrogio2015optimal}.

This paper focuses on the Wasserstein metric for three primary reasons. First, it offers an intuitive interpretation as an ``earth-moving" distance~\citep{rubner1998metric}, characterized by the cost function $c(x_1, x_2) = \frac{1}{2}\|x_1 - x_2\|^2$. Second, existing literature underscores the Wasserstein metric's role in imparting a rich geometric structure to the space of probabilities~\citep{mallasto2017learning,takatsu2011wasserstein,villani2021topics}. This geometric framework proves particularly advantageous in several domains, such as distributionally robust optimization, wherein Wasserstein balls are pivotal~\citep{blanchet2019robust,duchi2023distributionally,esfahani2015data,gao2023distributionally,luo2019decomposition,pflug2007ambiguity,rahimian2022frameworks,sinha2017certifying}. Specifically, this work emphasizes the gradient flows induced by the Wasserstein metric to facilitate flow matching, a concept elaborated in Section~\ref{sec:WGF}. The metric's unique capacity to yield straightforward and compelling applications through gradient flows is noteworthy~\citep{santambrogio2017euclidean}. However, establishing gradient flows, for instance, using Kullback-Leibler divergence within the framework of Bures-Wasserstein gradient flows, proves to be considerably less elegant mathematically~\citep{lambert2022variational}. Lastly, as illustrated in Remark~\ref{rmk:gaussian_W2}, the metric seamlessly integrates certain invariant properties a priori.

Further, by applying the Kantorovich-Rubinstein duality, we can derive the dual formulation of the 2-Wasserstein distance, as detailed in \textbf{Theorem~\ref{thm:W2_dual}}. This dual formulation offers well-established interpretations in the context of optimal transport theory~\citep{ambrosio2005gradient,panaretos2019statistical}.
\begin{theorem}[Dual representation; \citet{benamou2000computational}]\label{thm:W2_dual}
    The dual representation for $\gW_2$ is given by
    \begin{align*}
        &\gW_2^2(\mu,\nu) \\
        =& 2\sup_{u,v} \left\{ \int u\,d\mu + \int v\,d\nu \Bigg| u(x_1)+v(x_2)\leq c(x_1,x_2)\right\}.
    \end{align*}
\end{theorem}
The dual formulation enables the derivation of efficient and potentially closed-form solutions by employing standard methods from the field of convex optimization~\citep{finlay2020train}. For a deeper understanding of the underlying principles, see Remark~\ref{rmk:shipper} for intuitive insights.

\subsection{First Variation}
The Radon space, as defined on $\gP(\Omega)$, inherently supports representations in terms of density. Consequently, moving forward, we will represent each element within this space using probability densities $p$.

The first variation of a functional $\mathcal{F}$ on $\mathcal{P}(\Omega)$ is represented by a linear functional $\frac{\delta\mathcal{F}}{\delta p}$. This functional characterizes the response of $\mathcal{F}$ to any feasible perturbation $h = p_1 - p_2$, where $p_1, p_2 \in \mathcal{P}(\Omega)$. Mathematically, this is expressed as follows:
$$
\left. \frac  {d}{dt }\gF(p+t h)\right|_{{t =0}} = \int \frac{\delta\gF}{\delta p}(p)h(z) \, dz,
$$ 
with $t$ being a perturbation parameter. Essentially, this definition serves as a linear approximation, capturing how $\mathcal{F}$ changes in response to minor alterations in its arguments.

It is important to note that the first variation is fundamentally the Gateaux derivative of the functional. By setting this first variation to zero, one essentially identifies the stationary points of the functional. This process is analogous to finding points where the gradient vanishes in standard calculus~\citep{jordan1998variational}. Techniques involving the first variation, such as the use of influence functions in causal inference, are prevalent in various mathematical and statistical applications~\citep{cai2023c-learner,kennedy2022semiparametric,tsiatis2006semiparametric,van2006targeted}.

\subsection{Wasserstein Gradient Flows}\label{sec:WGF}
Our focus is on identifying a trajectory through the space of probability measures, $\mathcal{P}(\Omega)$, that minimizes a functional $\mathcal{F}: \mathcal{P}(\Omega) \to \mathbb{R}$. This functional $\mathcal{F}$ is characterized by its lower semi-continuity and boundedness from below. Additionally, the trajectory must maintain a certain level of smoothness. In this context, the Wasserstein gradient flow provides a framework for understanding the dynamics of probability measures. It describes how these measures evolve along the path of steepest descent with respect to the functional $\mathcal{F}$, guided by the Riemannian metric imparted by the 2-Wasserstein metric, as discussed in \citet{fan2021variational} in their variational approach.

Intuitively, this process involves iteratively updating the probability measure at each step $n$ in the space $\mathcal{P}(\Omega)$, as expressed in the following equation:
\begin{equation}\label{eq:grad_flow_min}
    p^{(n+1)} \in \argmin_{p} \gF(p) + \frac{\gW_2^2(p,p^{(n)})}{2\tau},
\end{equation}
while taking the limit $\tau\to 0$. As pointed out by \citet{de1993new,massari1999generalized}, this discrete formulation is consistent with the regime of generalized minimizing movements, which is discussed in Remark~\ref{rmk:proximal}. The following theorem illustrates the underlying mathematical principles governing the gradient flow for the minimization problem in the induced probability space.
\begin{theorem}[\citet{jordan1998variational,villani2021topics}]\label{thm:wgf}
    The gradient flow that addresses the continuous-time minimization problem in Eq.~\eqref{eq:grad_flow_min} is described by the Fokker-Planck equation
    \begin{equation}\label{eq:fokker-planck}
        \frac{dp}{dt} = \nabla\cdot \left( p\nabla \frac{\delta\gF}{\delta p}\right),
    \end{equation}
    where $\frac{\delta\mathcal{F}}{\delta p}$ represents the first variation of $\gF$.
\end{theorem}
% \begin{proof}
%     Refer to the Appendix.
% \end{proof}
\textbf{Theorem~\ref{thm:wgf}} states that the evolution of probability measures (denoted by $p$) that most steeply decreases the functional $\gF$ can be described by the Fokker-Planck equation, a partial differential equation that characterizes how the probability density function changes over time. In particular, this is the JKO scheme~\citep{jordan1998variational} in establishing time-discrete approximations to weak solutions of diffusion equations like Eq.~\eqref{eq:fokker-planck}.

In subsequent sections, we elucidate that gradient flows within the probability space serve as a navigational tool, steering the evolution of a distribution from an initial state \(p_0\) towards the target data distribution \(p_1\). While various metrics and divergences can be employed to define these gradient flows, they often lack certain intuitive and geometric characteristics that render Wasserstein gradient flows particularly advantageous \citep{ambrosio2005gradient,ambrosio2013user,otto2001geometry,villani2009optimal}. Consequently, this paper opts for the renowned Wasserstein metric, celebrated for its profound geometric properties and its inherent connection to optimal transport. This metric facilitates a coherent comprehension of the manner in which distributions traverse the space of probability measures \citep{rubner1998metric}.

\section{NORMALIZING WASSERSTEIN FLOWS}
The evolution in \textbf{Theorem~\ref{thm:wgf}} naturally leads to the consideration of continuous normalizing flows (CNFs) as a powerful tool for modeling complex distributions~\citep{kobyzev2020normalizing}, which are a class of generative models that aim to transform a simple base distribution into a more intricate target distribution through a smooth and invertible mapping, as delineated in \citet{chen2018neural}. 

Given a neural ODE defined as
\begin{equation}\label{eq:neural_ODE}
    \frac{dz}{dt} = f(z,t;\theta),
\end{equation}
the change in log-density is governed by the differential equation representing the instantaneous change of variables:
\begin{equation}\label{eq:inst_change}
    \frac{\partial \log p(z(t))}{\partial t} = -\nabla \cdot f.
\end{equation}
Alternatively, even in the presence of a Fokker-Planck or Liouville PDE for the system's evolution, Eq.~\eqref{eq:inst_change} can still be recovered (as shown by \citet{chen2018neural}, Eq. (32)) by tracking the trajectory of a particle \( z(t) \) rather than examining \( p(\cdot, t) \) at a fixed point.

Following the approach of \citet{chen2018neural,chen2020learning,grathwohl2018ffjord}, continuous normalizing flows are trained to maximize the terminal log-likelihood at \( t_1 \), which is also a learnable parameter:
\begin{equation}
    \log p(z(t_1)) = \log p(z(t_0)) - \int_{t_0}^{t_1} \nabla\cdot\left(\frac{\partial f}{\partial z} \right) dt.
\end{equation}
This framework is numerically tractable thanks to the well-established adjoint method~\citep{pontryagin2018mathematical} and stochastic trace estimators~\citep{adams2018estimating,grathwohl2018ffjord, hutchinson1989stochastic}.

The following figures show two basic continuous normalizing flows, originating from a prior distribution at \( t=0 \) (typically a standard Gaussian), and evolving into a distribution parameterized by neural networks that aligns with the observed data.

The key aspect of CNFs is the use of neural networks to parameterize the velocity field in a differential equation, thereby enabling the model to learn complex transformations. In the realm of gradient flows, CNFs can be effectively utilized to approximate the dynamics described by the Fokker-Planck equation in Eq.~\eqref{eq:fokker-planck}. By leveraging the concept of the Wasserstein gradient flow, one can guide the CNFs to match the evolution of probability measures as governed by the functional $\mathcal{F}$. We term such flows as the \emph{normalizing Wasserstein flows}. To achieve this, we propose adapting the CNFs to explicitly mimic the Wasserstein gradient flow, which involves two main components:
\begin{itemize}
    \item Optimizing the CNF parameters to minimize the discrepancy between the modeled distribution and the target distribution.
    \item Ensuring the invertibility and smoothness of the CNF, which is critical for maintaining the properties of the probability measures throughout the normalizing Wasserstein flows. 
\end{itemize}

The initial goal of aligning the learned probability density with the data distribution can be effectively accomplished, for instance, through maximum likelihood estimation. In this process, the probability density generated by the latent variables $z$ is progressively trained to closely approximate the data distribution. This alignment ensures that the model captures the underlying structure and variability present in the data.

However, the second objective diverges from simply focusing on the behavior of the learned probability density. Instead, it concentrates on aligning the gradients of the density evolutions. This concept forms the crux of diffusion models, which emphasize the alignment of the neural network with the score (gradient) of the probability distributions. In certain structured scenarios, this alignment is achieved through well-defined stochastic interpolants~\citep{albergo2022building,albergo2023stochastic1,albergo2023stochastic2}. What becomes paramount in these cases is not the density itself, but the gradient flow matching~\citep{lipman2022flow}, which is central to the second objective.

In the context of stochastically interpolating between the noisy prior $p_0$ and the data distribution $p_1$, the importance of this approach becomes evident in scenarios where there is an inclination to edge closer to the prior distribution, perceived as the ``truth", and move slightly away from a potentially biased data distribution. This shift can be particularly relevant in fields like causal inference or semiparametric literature, where understanding and approximating the underlying truth is crucial.
In such cases, using an interpolated point between the prior and data distributions as a proxy becomes a strategic approach. This method enables researchers to balance the innate biases present in the data with the theoretical constructs of the prior distribution. By judiciously selecting a point on the continuum between 
$p_0$ and $p_1$, one can effectively create a model that is not overly influenced by the idiosyncrasies of the observed data, while still retaining a connection to the empirical reality that the data represents~\citep{kennedy2022semiparametric,van2006targeted}.

Stochastic interpolation thus fulfills a dual role. First, it acts as a vital mechanism to counteract data biases, adeptly integrating the theoretical insights gleaned from the prior distribution. This integration is key in ensuring that the final analysis is not overly skewed by anomalies or specific trends in the dataset. Second, it enables a more sophisticated exploration of the data generation process, grounded in theory. This exploration goes beyond surface-level observations, delving into the deeper, underlying mechanisms that drive the data.

In this context, the potential limitations in the expressive power of neural ODEs do not restrict the extensive applicability of normalizing Wasserstein flows. Two primary considerations support this assertion. First, numerous recent advancements, such as Augmented Neural ODEs, have expanded the class of functions modelled by neural networks to encompass all diffeomorphisms~\citep{dupont2019augmented, zhang2019approximation}. Second, the framework of normalizing Wasserstein flows is equally applicable to stochastic differential equations (SDEs), including Langevin flows, as demonstrated in various studies~\citep{chen2018continuous, jankowiak2018pathwise, rezende2015variational, salimans2015markov, welling2011bayesian}. Further details on this topic will be elaborated in Section~\ref{sec:interpretations}.

Building on these principles, we propose a methodology for achieving geometry-aware interpolation. This method is designed to optimize causal inference, or in the broader context of statistical functional estimation, adhering to the principles of semiparametric efficiency. By incorporating geometrical considerations into the interpolation process, we aim to refine our understanding of the data's structure and causal relationships. This approach not only aligns with the theoretical framework but also enhances the practical effectiveness of causal inference, ensuring that the conclusions drawn are both statistically sound and relevant to the real-world phenomena they aim to explain.

% In the following, we demonstrate two examples of practical concerns. The first example incorporates variance regularization in the continuous normalizing flows via the corresponding form of Wasserstein gradient flows. The second generalizes to semiparametric robust estimation and causal inference.

\subsection{Motivating Example: Variance Regularization}

In the pursuit of learning a smooth distribution, one approach involves controlling its variance or its surrogates~\citep{namkoong2017variance}. This is achievable by converting the distribution into a standard normal form using continuous normalizing flows and adding a penalty on its variance. We introduce a novel objective that integrates a variance regularization term with the initial loss function, aiming to control the distribution's variance. We present a theorem describing the evolution of the log density under the associated Wasserstein gradient flow. 

\begin{theorem}\label{thm:var_reg}
    Operating on the space of zero-mean densities, i.e.,
    \[
    \int_{\mathbb{R}^n} z p(z) \, dz = 0,
    \]
    the Wasserstein gradient flow, aimed at minimizing the variance functional
    \[
    \gF(p) = \frac{1}{2}\int_{\mathbb{R}^n} z^2 p(z) \, dz,
    \]
    is described by the partial differential equation
    % \[
    % \frac{\partial \log p}{\partial t} =  \frac{\nabla p}{p} \cdot z + n.
    % \]
$$
\frac{\partial \log p}{\partial t} = \1 + (\nabla \cdot \log p ) z.
$$
\end{theorem}
In the course of our analysis, it becomes evident that the drift component in the Fokker-Planck equation exhibits a direct proportionality to the variable $z$. This observation implies a temporal dynamic where the probability mass is increasingly centralized around zero. The lack of a diffusion term further accentuates this behavior, as it precludes the dispersion of the probability mass, thereby catalyzing the convergence to a Dirac delta distribution at zero. Consequently, this dynamic inherently favors probability distributions characterized by diminishing variance over time.

To mitigate potential mode collapse, a phenomenon where the distribution narrows down to a point mass~\citep{che2016mode,salimans2016improved}, our approach integrates actual data into the optimization objective. Drawing inspiration from the domain of physics-informed neural networks~\citep{cai2021physics,karniadakis2021physics}, we incorporate a variance regularization term into the loss function. The modified loss function $\tilde{\mathcal{L}}$, parametrized by $\theta$ and evaluated at time $t_1$, is formulated as:
\begin{align*}
    \tilde{\mathcal{L}}(\theta, t_1; z) &= \mathcal{L}(z(t_1)) \\
    +& \lambda \int_0^1 \mathbb{E}\left\| \frac{\partial \log p(z)}{\partial t} - \1 - (\nabla \cdot \log p ) z \right\|_{\ell_2}^2 \, dt,
\end{align*}
where $\lambda$ represents the regularization coefficient. This formulation intuitively utilizes the divergence of the log-likelihood, indicative of regions with concentration or dispersion of the probability density, as a means to regularize the variance.

It should be noted that the concept of regularization within the framework of normalizing flows is not novel. Various approaches have been explored, targeting domain adaptation. This includes applications in graphical settings \citep{wehenkel2021graphical} and image classification tasks \citep{izmailov2020semi}, as well as the superresolution of imaging \citep{altekruger2022patchnr}. Additionally, just like regularization restricts the function class in general~\citep{hou2022spectral}, there are also approaches that impose regularity constraints on the class of functions learned by the flows, such as ensuring smoothness~\citep{kohler2021smooth}, geodesic constraints~\citep{salman2018deep}, and uniqueness of the solutions~\citep{finlay2020train}. Notably, the latter addresses the issue of non-uniqueness in the vector field (for example, as learned from FFJORD \citep{grathwohl2018ffjord}), by enforcing the solution trajectories to adhere to straight lines with constant velocity. In the context of normalizing Wasserstein flows, we extend this concept by generalizing the solution trajectories to be more geometrically attuned to other external objectives.

While the approach of normalizing Wasserstein flows effectively employs the PDE residual loss as a regularization term, it introduces a complexity: the need for gradients of the log-likelihood relative to the latent variables $z$. This requirement can be computationally challenging or necessitate the use of finite difference methods and numerical gradients, as opposed to straightforward gradient updates on the parameter vector $\theta$. In the subsequent section, we propose a generalized framework that circumvents this complexity by focusing solely on the alignment of gradients, or what may be termed as ``velocity fields".

\subsection{General Formulation}
Inspired by Appendix A2 of \cite{chen2018neural}, and recent developments in velocity-based diffusion models \cite{albergo2023stochastic1}, to simplify computations, Eq.~\eqref{eq:inst_change} suggests focusing the regularization on the velocity field $-f$ in the loss function. This is supported by the following theorem.

\begin{theorem}\label{thm:general_formulation}
    Suppose the Liouville equation characterizes the density evolution at a fixed point $z^*$ as follows:
    \[
    \frac{\partial p(z^*,t)}{\partial t} = \nabla \cdot \left(p(z^*,t)\nabla \frac{\delta\mathcal{F}}{\delta p}(z^*,t)\right).
    \]
    The conditions of Eq.~\eqref{eq:neural_ODE} and \eqref{eq:inst_change} in the neural ODE framework are satisfied when the following velocity field alignment holds:
\begin{equation}\label{eq:velocity_alignment}
        \frac{dz}{dt} + \nabla \frac{\delta\mathcal{F}}{\delta p} = 0.
\end{equation}
\end{theorem}
This theorem ensures proximity to $\nabla \frac{\delta\mathcal{F}}{\delta P}$. This formulation incorporates the geometry induced by a generic functional $\mathcal{F}$, while also maximizing the likelihood of observed data. Accordingly, Eq.~\eqref{eq:velocity_alignment} yields a natural loss function defined as:
\begin{equation}\label{eq:regularized_loss}
    \tilde{\mathcal{L}}(\theta, t_1; z) = \mathcal{L}(z(t_1)) + \lambda \int_0^1 \mathbb{E}\left\| \frac{dz}{dt} + \nabla \frac{\delta\mathcal{F}}{\delta p} \right\|_{\ell_2}^2 \, dt.
\end{equation}

When $\mathcal{F}$ is the variance functional, the density evolution from $t_0$ to $t_1$ aims at minimizing variance. Conversely, when moving backward in time, the prior density (e.g., a standard normal distribution) evolves towards maximizing variance while maintaining a high likelihood.

Remarkably, the framework we propose here does not only work for variance regularization. Alternative geometry could be induced as well, such as regularization based on expectations or negative entropy, where the latter is often well-motivated by the heat equation, as shown in Remark~\ref{rmk:entropy}. 

\subsection{Interpretations}\label{sec:interpretations}
This section elucidates two distinct interpretations of the loss function as defined in Eq.~\eqref{eq:regularized_loss}, aiming to elucidate the inherent trade-offs. These interpretations will serve as a foundation for more in-depth analysis in future works.

In a more formal exposition, consider a stochastic interpolant $X_t \sim p_t$ that smoothly transitions between $X_0 \sim p_0$ and $X_1 \sim p_1$, adhering to the requisite boundary conditions~\citep{albergo2023stochastic1}.
The interpolant is defined as:
\begin{equation}\label{eq:stochastic_interpolant}
    X_t = I(t,X_0,X_1) + \gamma(t)\epsilon
\end{equation}
where $\epsilon$ represents a standard Gaussian noise, assumed to be independent of all other variables. To alleviate notational complexity, we set $t_0=0$ and $t_1=1$ without loss of generality.

\subsubsection{Optimal Transport Interpretation} 
This interpretation principally engages with optimal transport theory. In the limit as the regularization coefficient $\lambda \to \infty$, the model asymptotically approaches a pure optimal transport framework. This convergence embodies the geometric principles underpinning Wasserstein gradient flows, as explicated in \citep{villani2021topics}.
A central element of this interpretation is the incorporation of the Benamou–Brenier formulation, which conceptualizes the optimal transport problem via velocity fields \citep{benamou2000computational}.

\begin{theorem}\label{thm:lower_bound_reg_loss}
The 2-Wasserstein metric between the initial distribution $p_0$ and the terminal distribution $p_1$ provides a sharp lower bound on the regularization term in Eq.~\eqref{eq:regularized_loss}:
$$
    \int_0^1 \E \left\|\frac{dz}{dt} +\nabla \frac{\delta\mathcal{F}}{\delta p}\right\|_{\ell_2}^2 \, dt \geq \gW_2^2(p_0,p_1).
$$
\end{theorem}

This formulation of the regularization term, in its rudimentary form, seeks a balance between the intrinsic data structure and an external objective which re-defines the notion of optimality when transporting from the initial distribution to the target one. The intrinsic data is further encapsulated in the first term $\mathcal{L}(z(t_1))$ through approaches such as maximum likelihood estimation, which can be reinterpreted as an optimal transport objective but with an alternative cost function. Such a function might incorporate structured priors \citep{alvarez2018structured} or domain-specific knowledge \citep{chuang2023infoot}. Importantly, when considering the geometry induced by $\mathcal{L}(z(t_1))$, assuming convexity is reasonable. This assumption is underpinned by constraining the normalizing flows to a class of functions that exclusively parameterize convex functions via neural networks \citep{alvarez2021optimizing}. Additionally, this term can be viewed as a disruptor to the pure Wasserstein geometry, as posited by the Kantorovitch formulation. This disruption might be analogous to the Gromov-Wasserstein geometry \citep{sturm2006geometry} or its regularizers \citep{vayer2018fused}, which juxtapose distributions along normalizing Wasserstein flows against actual data distributions situated in disparate metric spaces.

\subsubsection{Diffusion Equilibrium Interpretation}
The second interpretation relates to the stationary distributions of two stochastic processes. The first process is characterized by ODE neural networks or diffusion models, tailored to empirical data as reflected in $\mathcal{L}(z(t_1))$. The second process adheres to flows or diffusions that precisely respect the Wasserstein geometry. In essence, the objective seeks to synergize the minimizers of these divergent objectives \citep{albergo2023stochastic1}, fostering a balance between empirical data fidelity and geometric congruence.

In the context of Eq.~\eqref{eq:stochastic_interpolant}, the gradient of the first variation $\nabla \frac{\delta\mathcal{F}}{\delta p}$ is identified as the unique minimizer of a specific quadratic objective~\citep{albergo2023stochastic1}:
\begin{align*}
    \nabla \frac{\delta\mathcal{F}}{\delta p} 
    &= \argmin_v \int_0^1 \E\Bigg[ \frac{1}{2}\|v(t,X_t)\|^2 \\
    &+ \left( \frac{\partial} {\partial t} I(t,X_0,X_1) + \dot\gamma(t) \epsilon \right) \cdot v(t,X_t) \Bigg] dt.
\end{align*}
Analogously, the time derivative of the latent variable $z$, $\frac{dz}{dt}$, is also characterized as the unique minimizer of a related quadratic objective. The alignment between these two velocity fields, namely $\nabla \frac{\delta\mathcal{F}}{\delta p}$ and $\frac{dz}{dt}$, is achieved through a synthesis of the two quadratic objectives. This synthesis provides a variational foundation for Eq.~\eqref{eq:regularized_loss}, effectively bridging the conceptual gap between stochastic interpolants and the optimization problem at hand.

% In the following figure, we compare the normalizing flows without regularization (top), with geometry induced by a common parametric submodel (medium), and with geometry induced by variance (low). In particular, we follow the dataset of a mixture of 8 Gaussian distributions with equal variance and with the mean evenly located on a circle around the origin. Furthermore, we assume there is one Gaussian distribution among all eight that gets randomly censored, i.e., there is a chance that the data from that distribution is not labeled, and instead, we assume the censored points are present in the dataset as the mean of that specific distribution. To conduct counterfactual inference, the goal is to reconstruct those censored points in a model-agnostic way. Intuitively, with a relatively weak inductive bias on the density geometry that interpolates between the true distribution and the censored one, one may recover the missing or unlabeled data.

% In this setting, the standard normalizing flows tend to overfit the data, which puts much higher weight on the mean of the censored distribution, as it has a lot of point mass. However, the variance-induced geometry provides more guidance on the direction to transform the density. 

% We demonstrate in the Appendix that the continuous normalizing flows transform the standard normal distribution to more complicated empirical ones, but with different degrees of smoothness as we vary the strength of regularization.

\section{OPTIMAL CAUSAL INFERENCE VIA OPTIMAL TRANSPORT}
This section elaborates on the potential applications of causal inference, particularly focusing on addressing the disparity between sample and population distributions that often leads to biased estimation of population parameters. A typical scenario involves estimating the proportion of comments about specific underrepresented populations in a dataset where highly toxic comments might be censored, thereby never appearing in the dataset. This situation is further compounded by substantial distribution shifts, as identified by \citet{koh2021wilds}, introducing additional bias in the estimation process.

Traditional statistical methods have been developed to adjust for this selection bias, notably including Inverse-Propensity Weighting (IPW) \citep{rosenbaum1983central, rosenbaum1984reducing} and G-methods \citep{robins1986new}. However, these methods often suffer from inefficiencies and can lead to erroneous conclusions, particularly when the underlying parametric models are misspecified. In contrast, semiparametric efficiency can be more reliably achieved through methods like Targeted Maximum Likelihood Estimation (TMLE) \citep{van2006targeted, van2011targeted} and Augmented IPW (AIPW) \citep{robins1994estimation, robins1995semiparametric, scharfstein1999adjusting}. These doubly robust methods, enhanced by nuisance estimators from deep learning and machine learning \citep{chernozhukov2018double} and cross-fitting, offer an optimal bias-variance tradeoff for the target parameter in semiparametric or nonparametric models \citep{schuler2017targeted}. This tradeoff is asymptotically optimal as the variance of the target parameter estimate approaches the variance of the influence function \citep{kennedy2022semiparametric}, thus correcting the first-order bias.

Building upon the foundational concepts of optimal transport and Wasserstein gradient flows, as previously discussed, this section delves deeper into their application within the realm of causal inference. \textbf{Theorem~\ref{thm:general_formulation}}, as established earlier, proposes an innovative method of integrating normalizing flows with an optimally directed approach to achieve minimal variance in causal inference models. This section further explores the role of information geometry, particularly in enhancing the semiparametric Cramér-Rao lower bound. By harnessing the principles derived from optimal transport theory, we endeavor to develop causal inference methodologies that not only adhere to the principles of minimal variance in finite-sample settings but also compare favorably against traditional methods like TMLE and AIPW, which primarily emphasize asymptotic optimality.

\subsection{Target Functional and Parametric Submodels}

Consider a sequence of statistical models $\mathcal{M} = \{p_t\} \subset \mathcal{P}(\Omega)$ indexed by $t \in [0, \infty)$, where the score function $g$ is characterized by zero mean and bounded variance with respect to the true density $p_0$. This setup assumes Hellinger differentiability—a key regularity condition—restricted to a locally asymptotic normal family where the score function is well-defined:
\begin{equation}\label{eq:diff_quad_mean}
    \int \left(\frac{\sqrt{p_t} - \sqrt{p_0}}{t} - \frac{1}{2} g_0 \sqrt{p_0} \right)^2 dz \to 0 \text{ as } t \to 0.
\end{equation}

Targeted Maximum Likelihood Estimation (TMLE) identifies a least favorable parametric submodel~\citep{van2017generally} that encompasses the true density $p_0$~\citep{stein1956efficient}. This is achieved through an initial estimator of the data distribution $p_1$, subsequently refined to identify the optimal index within the parametric submodel:
\begin{equation*}
p_t = p_0(1 + tg_0).
\end{equation*}
Consequently, we derive that
\begin{equation*}
  g_0 = \left. \frac{\partial}{\partial t} \log p_t \right|_{t = 0}
\end{equation*}
is the score at the ground truth, a direct implication from Eq.~\eqref{eq:diff_quad_mean}, as demonstrated in Lemma 7.6 of \citet{van2000asymptotic}.

In the realm of causal inference, our interest lies in evaluating a \textit{target functional} $\psi: \mathcal{M} \to \mathbb{R}^d$ at the true measure $\psi(p_0)$. For instance, under the assumptions of positivity and censoring at random \citep{kennedy2022semiparametric}, the target functional for mean response can be defined as an expected regression functional
\begin{equation*}
\psi(p) = \mathbb{E}_p[Y] = \mathbb{E}_p\left[\mathbb{E}_p[Y \mid A = 1, X]\right]
\end{equation*}
evaluated at $p = p_0$.

\subsection{Local Semiparametric Efficiency}

To elucidate the first-order bias in a misspecified model $p_t$, we employ the von Mises expansion (or distributional Taylor expansion) at the true model $p_0$ \citep{fernholz2012mises}:
\begin{equation*}
    \psi(p_t) - \psi(p_0) = \int \varphi(p_0) (p_t - p_0) dx + R_2(p_t, p_0),
\end{equation*}
where $\varphi$ represents the efficient influence function.

Considering Neyman orthogonality \citep{chernozhukov2018double, chernozhukov2022locally}, and assuming pathwise differentiability of the target functional $\psi(p_t)$, we have:
\begin{equation*}
\frac{d}{dt} \psi(p_t) = \langle \varphi(p_t), g_t \rangle_{p_t},
\end{equation*}
with a generic score
\begin{equation}\label{eq:score}
    g_t = \frac{\partial}{\partial t} \log p_t \in \gT_{p_t}\gM.
\end{equation}
Thus, in estimating a statistical functional $\psi(p_t)$ at $t = 0$, the tightest semiparametric Cramér-Rao lower bound is expressed as:
\begin{align*}
    \sup_{g_0\in \gT_{p_0}\gM}\frac{\left(\left.\frac{d}{dt}\psi(p_t)\right|_{t=0}\right)^2}{\langle g_0, g_0 \rangle_{p_0}} &= \sup_{g_0\in\gT_{p_0}\gM} \frac{\langle \varphi(p_0), g_0 \rangle_{p_0}^2}{\langle g_0, g_0 \rangle_{p_0}} \\
    &\leq \langle \varphi(p_0), \varphi(p_0) \rangle_{p_0} \\
    &= \Var_{p_0}(\varphi(p_0)),
\end{align*}
where equality is achieved at $g_0 = \varphi(p_0) \in \mathcal{T}_{p_0}\mathcal{M}$, following the Cauchy–Schwarz inequality. 
Apparently, for $t=0$, we have the Cramér-Rao lower bound, which shows that the efficiency is achieved when the score function $g_0$ is equal to the efficient influence function $\varphi(p_0)$.
This congruence signifies an optimal estimation under certain regularity conditions. Nevertheless, as we extend our consideration to a general $t$, the task of identifying an efficient influence function $\varphi(p_t)$, such that it consistently resides within the tangent space $\mathcal{T}_{p_t}\mathcal{M}$ of the model manifold $\mathcal{M}$, becomes arduous. The complexity of this problem escalates due to the dynamic nature of the tangent space, which evolves as the model $p_t$ changes.

The score function $g_t$, defined in Eq.~\eqref{eq:score} as the derivative of the log-likelihood with respect to the parameter at $p_t$, captures only the \textit{local} sensitivity of the model at this point. In contrast, the efficient influence function $\varphi(p_t)$, apart from encapsulating this local sensitivity, also embodies the capacity to reflect the \textit{global} structural intricacies of the statistical model. This global perspective is critical in relation to the target functional $\psi$, which the efficient influence function aims to estimate.

Given this context, the concept of semiparametric efficiency appears inherently local. A pertinent query then emerges: Is it feasible to pursue global semiparametric efficiency? One promising approach is to minimize the semiparametric efficiency bound over the entire trajectory of the model manifold $\gM$. This minimization can be conceptualized through the framework of normalizing Wasserstein flows. Specifically, the Wasserstein gradient flows, designed to minimize the Cramér-Rao lower bound, incorporate considerations of the global structural attributes of the model. Formally, the objective is to minimize the functional:
$$
\gF(p_t) = \Var_{p_t}(\varphi(p_t)),
$$
where $\gF$ represents the variance of the efficient influence function under the distribution $p_t$. This minimization can be operationalized by formulating a velocity regularization term, as delineated in Eq.~\eqref{eq:regularized_loss}, which governs the evolution of the flows.

% So far, we have noted that the sharpest semiparametric efficiency bound is given by the variance of the efficient influence function, provided that Neyman orthogonality holds. It is also worth noting that with a different form of score function $g$, we may have a different efficient influence function $\varphi$ that is orthogonal to the score, the one with the highest variance yields the tightest Cramér-Rao lower bound. To achieve such flexibility in the representation of score functions, we have to relax Eq.~\eqref{eq:diff_quad_mean} to solving for some score functions that \textit{approximately} equals to the differential of the quadratic mean density; otherwise the score is also the derivate of log density w.r.t. the parameter (\citet{van2000asymptotic}, Lemma 7.6).

% Fortunately, the continuous Wasserstein flows allow us to \textit{approximately} solve Eq.~\eqref{eq:diff_quad_mean} in the direction that maximizes the Cramér-Rao lower bound. 
% Similar to before, we design the loss function
% $$
% \tilde\gL(\theta,t_1;z) = \gL(z(t_1)) + \lambda \E\left| \frac{\partial p}{\partial t} -\nabla\cdot \left( P\nabla \frac{\delta\gF}{\delta P}\right) \right|^2,
% $$
% where $\gF$ is the functional corresponding to the Cramér-Rao lower bound.

\subsection{Experiments}
Consider a simple illustrative example where our objective is to estimate the target functional corresponding to the marginal population mean of a two-dimensional random variable $(X, \cdot): \Omega \to \mathbb{R}^2$. The target functional in this context is defined as:
\begin{equation*}
\psi(p) = \mathbb{E}_p[X].
\end{equation*}
The efficient influence function can be straightforwardly derived as:
\begin{equation*}
\varphi(p) = X - \mathbb{E}_p[X].
\end{equation*}
In accordance with Eq.~\eqref{eq:regularized_loss}, we align the ODE neural network, represented by $f = \frac{dz}{dt}$, with the Wasserstein gradient flows. These flows are aimed at minimizing the functional $\gF = \Var_{p_t}(\varphi(p_t))$.

\begin{figure}[H]
      \centering
\includegraphics[width=0.5\textwidth]{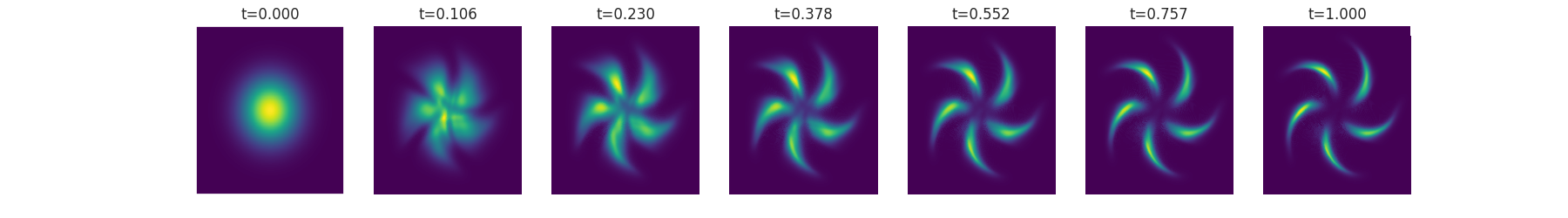}
\caption{Illustration of Trajectories in Continuous Normalizing Flows: This figure presents the evolution of continuous normalizing flows from a 2-dimensional standard Gaussian prior to distinct data distributions. Specifically, the trajectories are shown for transformations leading to the data distributions generated by two datasets: the \textsc{8Gaussians} dataset and the \textsc{Pinwheel} dataset, respectively.}
\label{fig:cnf}
\end{figure}

Suppose the true distribution $p_0$ is a standard Gaussian, which coincidentally also serves as the prior in continuous normalizing flows. The data distribution is characterized by $p_1$, as depicted in Figure~\ref{fig:cnf}. For instance, in the case of the \textsc{8gaussians} dataset, the marginal distribution of $X$ is gradually subjected to finite-sample bias, as illustrated in the subsequent figure.

With varying degrees of regularization strength as specified in Eq.~\eqref{eq:regularized_loss}, we derive the normalizing Wasserstein flows that effectively bridge the true distribution $p_0$ and the data distribution $p_1$. The subsequent figures visualize these flows for several toy datasets. Notably, the mean-squared error (MSE) associated with the normalizing Wasserstein flows is consistently lower than that of the naive flows, which lack regularization, at all points along the trajectory. This observation is crucial, as the finite-sample estimation error is fundamentally characterized by the MSE of plug-in estimators. Consequently, our analysis demonstrates the feasibility of more efficient estimators in finite-sample scenarios. Such estimators facilitate optimal causal inference, optimally positioned within the trajectory connecting $p_0$ and $p_1$.

\begin{figure}[H]
        \centering
        \includegraphics[width=0.45\textwidth]{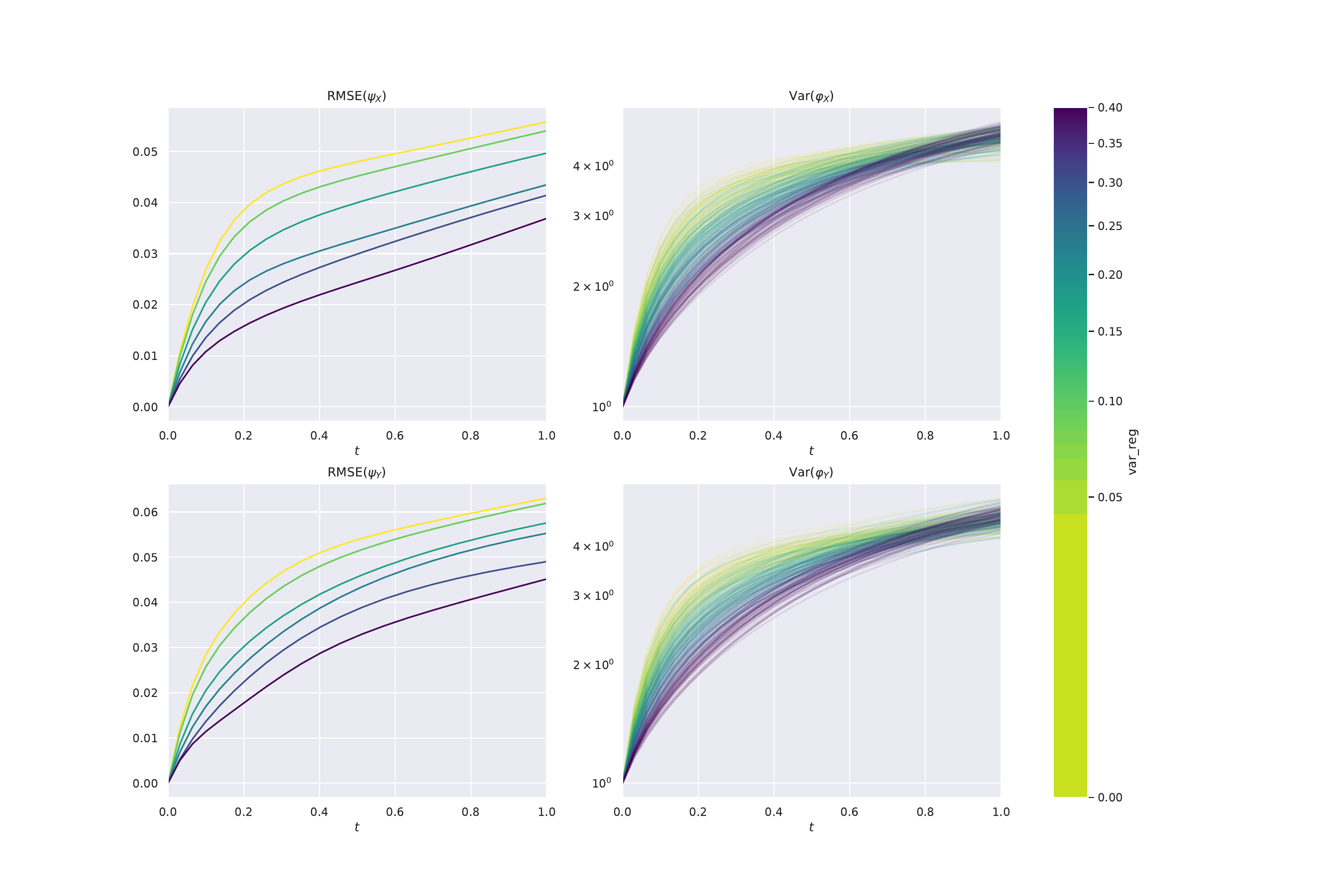} 
        \caption{Analysis of Normalizing Wasserstein Flows on \textsc{8gaussians}}
        \label{fig:8gaussians}
\end{figure}

Figure~\ref{fig:8gaussians} presents a dual analysis. The left panel illustrates the mean-squared error of plug-in estimators along the flow path, while the right panel displays the variance of the efficient influence function. It is evident that each flow attains semiparametric efficiency at the true distribution $p_0$. Notably, estimators with stronger velocity regularization demonstrate greater efficiency compared to those without any regularization.

\begin{figure}[H]
        \centering
        \includegraphics[width=0.45\textwidth]{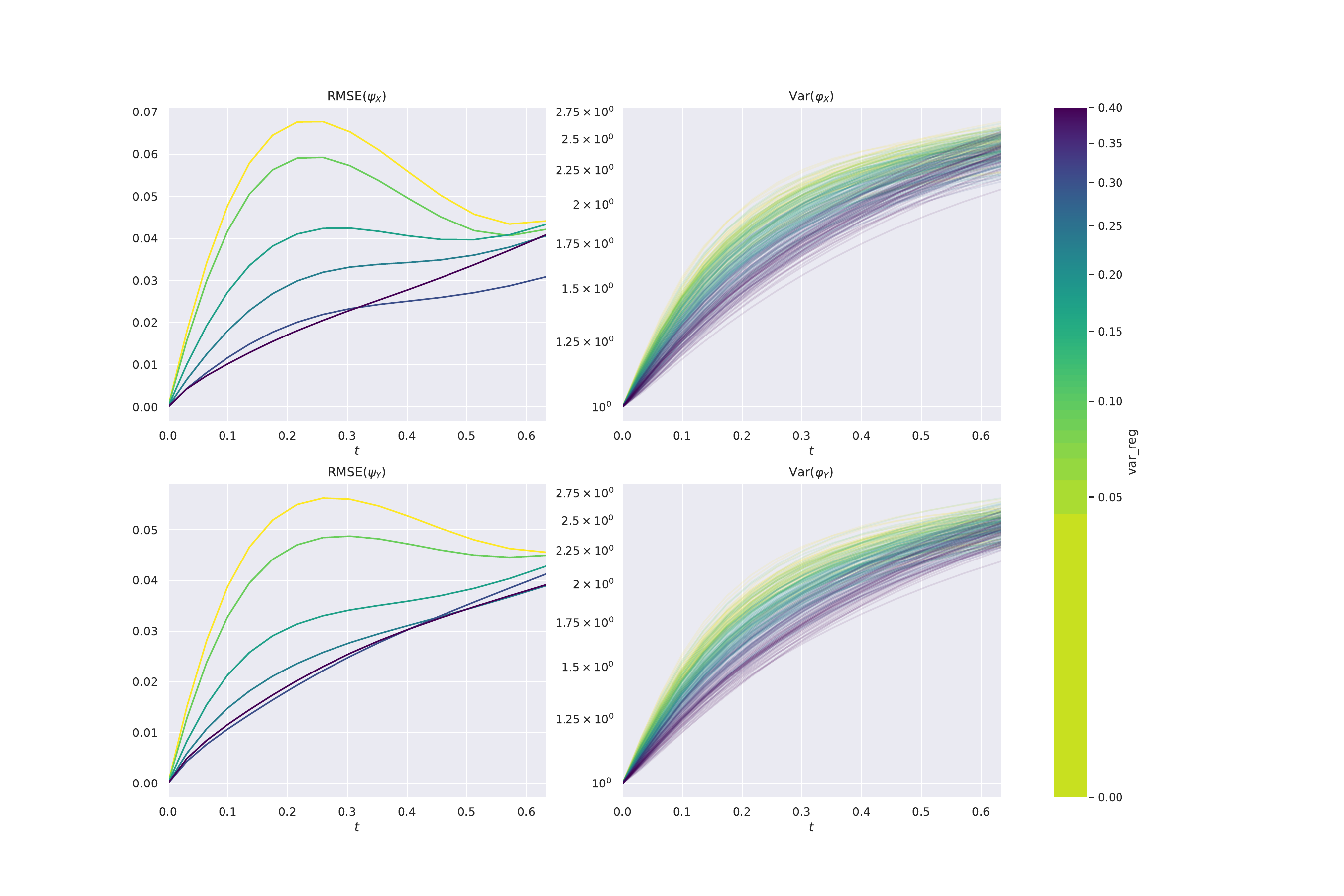} 
        \caption{Consistent Outcomes in Normalizing Wasserstein Flows for \textsc{pinwheel}}
        \label{fig:pinwheel}
\end{figure}

The results in Figure~\ref{fig:pinwheel} mirror those observed for the \textsc{8gaussians} dataset, showcasing similar patterns in the efficiency and variance metrics for the studied flows.

\section{CONCLUSIONS}
In the conclusion, we summarize key findings and insights from our study.

First, TMLE effectively conducts interpolation between \( p_0 \) and \( p_1 \) by optimally fitting the index \( \epsilon \), achieving a state of zero bias. This method demonstrates precision in balancing the distributions, aligning closely with the theoretical ideal.

Second, our preliminary experimental results indicate that the RMSE of estimators based on interpolation could universally be lower than that of TMLE. This potential improvement may be attributed to the allowance for some level of bias, which, counterintuitively, can enhance overall accuracy.

Lastly, when equipped with oracle Riesz representers, which are the correctly specified perturbation directions of an initial statistical model, both TMLE and the C-Learner can precisely project onto a space where the efficient influence function has a zero empirical mean. This leads to asymptotic unbiasedness. In contrast, our approach involves an approximate projection of the data-inspired statistical model along a trajectory. This trajectory, which we define as the geometry-aware normalizing Wasserstein flows, is designed to attain semiparametric efficiency almost ubiquitously. It represents a sophisticated path that integrates geometrical insights with statistical efficiency, offering a nuanced and theoretically grounded approach to modeling and inference.

% \section*{Acknowledgements}

% \textbf{Do not} include acknowledgements in the initial version of
% the paper submitted for blind review.

% If a paper is accepted, the final camera-ready version can (and
% probably should) include acknowledgements. In this case, please
% place such acknowledgements in an unnumbered section at the
% end of the paper. Typically, this will include thanks to reviewers
% who gave useful comments, to colleagues who contributed to the ideas,
% and to funding agencies and corporate sponsors that provided financial


\bibliography{literature}
\bibliographystyle{ICML2024/icml2024}

%%%%%%%%%%%%%%%%%%%%%%%%%%%%%%%%%%%%%%%%%%%%%%%%%%%%%%%%%%%%%%%%%%%%%%%%%%%%%%%
%%%%%%%%%%%%%%%%%%%%%%%%%%%%%%%%%%%%%%%%%%%%%%%%%%%%%%%%%%%%%%%%%%%%%%%%%%%%%%%
% APPENDIX
%%%%%%%%%%%%%%%%%%%%%%%%%%%%%%%%%%%%%%%%%%%%%%%%%%%%%%%%%%%%%%%%%%%%%%%%%%%%%%%
%%%%%%%%%%%%%%%%%%%%%%%%%%%%%%%%%%%%%%%%%%%%%%%%%%%%%%%%%%%%%%%%%%%%%%%%%%%%%%%
\newpage
\appendix
\onecolumn
\section{Proofs}
\begin{proof}[Proof of Theorem~\ref{thm:W2_dual}]
Observe that
\begin{align*}
    \sup_{u,v}\int u\,d\mu + \int v\,d\nu - \int (u+v)\,d\gamma &= \begin{cases}
    0 & \gamma\in\Gamma(\mu,\nu) \\
    \infty & \text{o.w.}
\end{cases} \\
\inf_{\gamma\geq 0} \int c - (u+v) \,d\gamma &= \begin{cases}
    0 & u(x_1)+v(x_2)\leq c(x_1,x_2) \ \forall x_1,x_2 \\
    -\infty & \text{o.w.}
\end{cases}
\end{align*}
Then for a transport cost function $c(x_1,x_2)$:
\begin{align*}
    \inf_{\gamma\in\Gamma(\mu,\nu)} \int c(x_1,x_2) \,d\gamma &= \inf_{\gamma\geq 0} \left\{ \int c(x_1,x_2) \, d\gamma + \sup_{u,v}  \int u\,d\mu + \int v\,d\nu - \int (u+v)\,d\gamma \right\} \\
    &= \sup_{u,v} \left\{ \int u\,d\mu + \int v\,d\nu  + \inf_{\gamma\geq 0} \int c - (u+v) \,d\gamma \right\} \\
    &= \sup_{u,v} \left\{ \int u\,d\mu + \int v\,d\nu  \Bigg| u(x_1)+v(x_2)\leq c(x_1,x_2) \ \forall x_1, x_2 \right\}.
\end{align*}
\end{proof}

\begin{proof}[Proof of Theorem~\ref{thm:var_reg}]
Observe that the functional $\mathcal{F}(p)$ is a linear functional. Moreover, it is bounded from below by 0, i.e., $\mathcal{F}(p) \geq 0$ for all permissible values of $p$. Consider any feasible perturbation $h = p_1 - p_2$ where $p_1, p_2 \in \gP(\Omega)$. It is evident that
    \[
    \int_{\mathbb{R}^n} h \, dz = \int_{\mathbb{R}^n} p_1 \, dz - \int_{\mathbb{R}^n} p_2 \, dz = 0.
    \]
    We compute the first variation of $\gF$ by identifying $\frac{\delta \gF}{\delta p}$ such that for any feasible $h$:
    \begin{align*}
    \int_{\mathbb{R}^n} \frac{\delta \gF}{\delta p}(p)h \, dz &= \left. \frac{d}{dt} \gF(p + t h) \right|_{t=0} \\
    &= \left. \frac{d}{dt} \frac{1}{2}\int_{\mathbb{R}^n} z^2 (p + th) \, dz \right|_{t=0} \\
    &= \frac{1}{2}\int_{\mathbb{R}^n} z^2 h \, dz,
    \end{align*}
    or
    \[
    \frac{\delta \gF}{\delta p}(p(z)) = \frac{1}{2}z^2,
    \]
    and its gradient
    \[
    \nabla \frac{\delta \gF}{\delta p}(p(z)) = z.
    \]
    The result then follows by substituting into the Fokker-Planck equation in Eq.~\eqref{eq:fokker-planck}:
    \begin{align*}
\frac{\partial p(z)}{\partial t} &= \nabla\cdot \left( p\nabla \frac{\delta\gF}{\delta p}\right) = \nabla\cdot (pz) \\
&= p\nabla \cdot z + z \nabla \cdot p = p \1 + (\nabla \cdot p ) z,
    \end{align*}
    and applying the partial derivatives to the log-likelihood.
\end{proof}

\begin{proof}[Proof of Theorem~\ref{thm:general_formulation}]
    By tracking the particle trajectory \( z(t) \) instead of evaluating \( p \) at \( z^* \) and applying the chain rule, we obtain:
    \begin{align*}
&\frac{\partial p(z(t),t)}{\partial t} = p_1(z(t),t)\frac{\partial z(t)}{\partial t} + p_2(z(t),t) \\
&= \sum_{i} \frac{\partial p(z(t),t)}{\partial z_i(t)}\frac{\partial z(t)}{\partial t} + p(z(t),t)\nabla\cdot \nabla \frac{\delta\mathcal{F}}{\delta p}(z(t),t) \\
&+ \sum_{i} \frac{\partial p(z(t),t)}{\partial z_i(t)} \nabla_i \frac{\delta\mathcal{F}}{\delta p}(z(t),t) \\
&= p(z(t),t)\nabla\cdot \nabla \frac{\delta\mathcal{F}}{\delta p}(z(t),t) \\
&= p(z(t),t)\left( -\nabla \cdot f(z(t),t) \right),
    \end{align*}
   leading to the equation:
    \[
    \frac{\partial \log p(z(t))}{\partial t} -= -\nabla \cdot f.
    \]
\end{proof}

\begin{proof}[Proof of Theorem~\ref{thm:lower_bound_reg_loss}]
    Let $p(t,z)$ denote the density function of $I(t,X_0,X_1)$ in Eq.~\eqref{eq:stochastic_interpolant}, or equivalently, the density of $X_t$ when the diffusion term is $\gamma(t)=0$. The boundary conditions are established as:
$$
    p(0,z) = p_0(z), \quad
    p(1,z) = p_1(z).
$$
Introduce the Lagrangian coordinates $\mX(t,z)$, defined by the relation:
$$
    \frac{\partial \mX}{\partial t}(t,z) = -\left( \frac{dz}{dt} +\nabla \frac{\delta\mathcal{F}}{\delta p}\right)(t,\mX(t,z)), \quad \mX(0,z)=z.
$$
Here, mass conservation is articulated through the pushforward equation:
\begin{equation}\label{eq:pushforward}
\mX(t,\cdot)_\# p_0 = p(t,\cdot),
\end{equation}
which establishes a feasible transport map. In this context, ``feasibility" implies that
\begin{align*}
    &\gW_2^2(p_0,p_1) \\
    &\leq \int_{\R^n} p_0(z)\|\mX(1,z)-z\|_{\ell_2}^2 \, dz \\
    &= \int_{\R^n} p_0(z)\|\mX(1,z)-\mX(0,z)\|_{\ell_2}^2 \, dz \\
    &= \int_{\R^n} p_0(z)\left\|\int_0^1 \frac{\partial \mX}{\partial t}(t,z) \, dt\right\|_{\ell_2}^2 \, dz \\
    &\leq \int_{\R^n} \int_0^1 p_0(z) \left\|\frac{\partial \mX}{\partial t}(t,z) \right\|_{\ell_2}^2 \, dt \, dz \\
    &= \int_0^1 \int_{\R^n} p_0(z)\left\|\left(\frac{dz}{dt} +\nabla \frac{\delta\mathcal{F}}{\delta p}\right)(t,\mX(t,z))\right\|_{\ell_2}^2 \, dz \, dt \\
    &= \int_0^1 \int_{\R^n} p(t,z)\left\|\left(\frac{dz}{dt} +\nabla \frac{\delta\mathcal{F}}{\delta p}\right)(t,z)\right\|_{\ell_2}^2 \, dz \, dt \\
    &= \int_0^1 \E \left\|\frac{dz}{dt}+\nabla \frac{\delta\mathcal{F}}{\delta p}\right\|_{\ell_2}^2 \, dt,
\end{align*}
where the penultimate equality is derived from Eq.~\eqref{eq:pushforward}. It is important to note that the equalities from above are attainable~\citep{benamou2000computational} under the condition
$$
\frac{dz}{dt} = (\mI-\mT)\left( (1-t)\mI + t\mT \right)^{-1} - \nabla \frac{\delta\mathcal{F}}{\delta p}.
$$
Here, $\mX(1,z)=\mT(z)$ represents the optimal transport map facilitating the transition from the initial distribution $p_0$ to the target distribution $p_1$.
\end{proof}
%%%%%%%%%%%%%%%%%%%%%%%%%%%%%%%%%%%%%%%%%%%%%%%%%%%%%%%%%%%%%%%%%%%%%%%%%%%%%%%
%%%%%%%%%%%%%%%%%%%%%%%%%%%%%%%%%%%%%%%%%%%%%%%%%%%%%%%%%%%%%%%%%%%%%%%%%%%%%%%
\section{Remarks}
\begin{remark}\label{rmk:gaussian_W2}
Similar to the Kullback-Leibler divergence, the 2-Wasserstein metric depends exclusively on the moments when the distributions are uniquely indexed by their moments. Consider, for instance, two Gaussian distributions, $\mu \sim \mathcal{N}(m_1, \Sigma_1)$ and $\nu \sim \mathcal{N}(m_2, \Sigma_2)$. The squared 2-Wasserstein distance between $\mu$ and $\nu$ is given by:
$$
\gW_2^2(\mu,\nu) = \|m_1-m_2\|^2 + \Tr\left( \Sigma_1 + \Sigma_2 -2(\Sigma_1\Sigma_2)^\frac{1}{2}\right).
$$

This allows us to explore certain invariant properties in the distributions. For instance, if the covariance matrices are equal, i.e., $\Sigma_1 = \Sigma_2$, this simplifies the 2-Wasserstein distance to
$$
\gW_2(\mu,\nu) = \|m_1-m_2\|,
$$
highlighting that under such conditions, the distance depends solely on the difference between the mean vectors of the two distributions.
\end{remark}

\begin{remark}\label{rmk:shipper}
The following example of the shipper's problem from \citet{villani2021topics} offers an intuitive explanation of the dual representation of the Wasserstein metric.
    Suppose you are transferring a huge amount of coal from your mines to your factories. You can hire trucks to do this transportation problem, but you have to pay~$c(x_1, x_2)$ for each ton of coal transported from place $x_1$ to place $x_2$. As you are trying to \textbf{minimize the price you have to pay}, a shipper comes to you and tells you \emph{``My friend, I will ship all your coal with my own trucks and you won’t have to care of what goes where. I will just set a price $u(x_1)$ for loading one ton of coal at place~$x_1$, and a price $v(x_2)$ for unloading it at destination $x_2$. I will set the prices in such a way that your financial interest will be to let me handle all your transportation! Indeed, you can check very easily that $\forall x_1, x_2: u(x_1)+v(x_2)\leq c(x_1,x_2)$ (I am even ready to give compensations for some places, in the form of negative prices)."} Of course you accept the deal. Now, what Kantorovich’s duality tells you is that if this shipper is clever enough, then he can arrange the prices in such a way that \textbf{you will pay him (almost) as much as you would have been ready to spend on hiring trucks}.
\end{remark}

\begin{remark}\label{rmk:proximal}
    In Euclidean space, the gradient descent algorithm is formulated as follows:
\begin{equation}\label{eq:gradient_descent}
        \Theta^{(n+1)} = \Theta^{(n)} - \tau \nabla F(\Theta^{(n)}),
\end{equation}
where $\Theta^{(n)}$ denotes the parameter vector at iteration $n$, $\tau$ is the step size or learning rate, and $\nabla F(\Theta^{(n)})$ represents the gradient of the function $F$ at $\Theta^{(n)}$.
It is noteworthy that gradient descent can also be expressed in a proximal formulation, which aligns with the minimal-movement regime. This is represented as:
$$
    \Theta^{(n+1)} \in \argmin_{\Theta} F(\Theta) + \frac{\left\|\Theta-\Theta^{(n)}\right\|_2^2}{2\tau},
$$
leading to an implicit update rule derived from the first-order condition:
\begin{equation}\label{eq:proximal_gradient_descent}
\Theta^{(n+1)} = \Theta^{(n)} - \tau \nabla F(\Theta^{(n+1)}).
\end{equation}
Interestingly, Eq.~\eqref{eq:gradient_descent} and Eq.~\eqref{eq:proximal_gradient_descent} converge in the limit of $\tau\to0$, resulting in the continuous-time gradient flow:
    $$
\frac{d\Theta}{dt}=-\nabla F(\Theta).
    $$
    Extending the concept of gradient descent to non-Euclidean spaces, specifically to the realm of optimal transport, leads us to the proximal Wasserstein gradient descent. This formulation is expressed as:
\begin{equation}\label{eq:proximal_Wasserstein}
p^{(n+1)} \in \argmin_{p} \gF(p) + \frac{\gW_2^2(p,p^{(n)})}{2\tau},     
\end{equation}
where $p^{(n)}$ denotes the distribution at iteration $n$, $\mathcal{F}(p)$ is the function being minimized, and $W_2^2(p, p^{(n)})$ represents the squared 2-Wasserstein distance between the distributions $p$ and $p^{(n)}$. Here, $\tau$ serves as the step size parameter. This formulation encapsulates the idea of minimizing a functional while simultaneously considering the geometry of the underlying space as captured by the Wasserstein metric. In the limit, the continuous-time gradient flow is constructed as in Eq.~\eqref{eq:fokker-planck}.
\end{remark}

\begin{remark}\label{rmk:entropy}
    Suppose we want to minimize the functional of negative entropy
    $$
\gF(p) = \int p(z) \log p(z) \, dz.
    $$
    To compute the first variation, notice that
    \begin{align*}
\left. \frac  {d}{dt}\gF(p+t h)\right|_{t =0} &= \left. \frac  {d}{dt} \int (p+t h) \log (p+t h) \, dz \right|_{t =0} \\
&= \int \left. \frac  {d}{dt} (p+t h) \log (p+t h) \, dz \right|_{t =0} \\
&= \int h \left. \log (p+t h)\right|_{t =0} + h \, dz \\
&= \int (\log p + 1) h \, dz.
    \end{align*}
    Then Eq.~\eqref{eq:fokker-planck} becomes the heat equation
    \begin{align*}
\frac{dp}{dt} &= \nabla\cdot \left( p\nabla \frac{\delta\gF}{\delta p}\right) \\
&= \nabla\cdot \left( p\nabla (\log p+1)\right) \\
&= \nabla\cdot \nabla p \\
&= \Delta p,
    \end{align*}
    which becomes the heat equation.
\end{remark}

\section{Figures}

\begin{figure}[H]
      \centering
\includegraphics[width=0.5\textwidth]{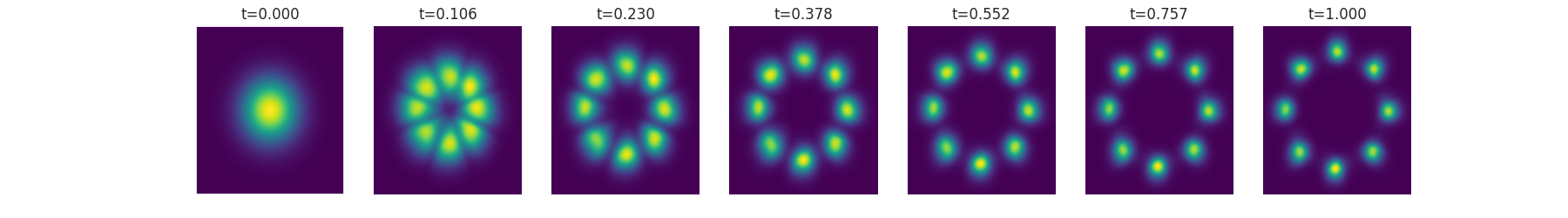}
\end{figure}

 \begin{figure}[H]
      \centering
                    \includegraphics[width=0.5\textwidth]{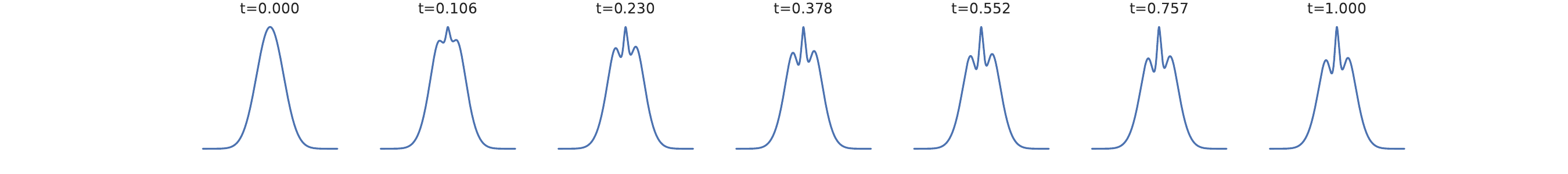}
                    \caption{Illustrative Transition from Initial Noise to Biased Sample Distribution: This figure depicts the evolution of the first dimension in the \textsc{8gaussians} dataset, starting from pure noise and gradually transitioning towards a biased sample distribution.}
\end{figure}

\end{document}